\RequirePackage{amsmath}
\RequirePackage{amssymb}
\documentclass[runningheads,a4paper]{llncs}
\usepackage{microtype}
\usepackage{bm}
\usepackage{xcolor}
\usepackage[spaces,hyphens]{url} 

\usepackage{cleveref}
\usepackage{booktabs}
\usepackage{tabularx}
\usepackage{graphicx}
\usepackage{amsmath}
\usepackage{amssymb}
\usepackage{ltl}
\usepackage{stackrel}
\usepackage{mathtools}
\usepackage{etoolbox}
\usepackage{makecell}
\usepackage{array}
\usepackage{todonotes}
\usepackage{enumitem}
\usepackage{comment}
\usepackage{float}
\newcolumntype{C}{>{\centering\arraybackslash}X}
\apptocmd{\sloppy}{\hbadness 10000\relax}{}{}

\graphicspath{ {./images/} }
\begin{document}
\title{Mining International Political Norms from the \\ GDELT Database}
%
%
\author{Rohit Murali\inst{1}\orcidID{0000-0002-6892-5830} \and
 Suravi Patnaik\inst{2}\and\\
 Stephen Cranefield\inst{3}\orcidID{0000-0001-5638-1648}}
\authorrunning{R. Murali et al.}
%
\institute{Indian Institute of Science, Bangalore, India \\ \email{muralirohit@gmail.com}\\
 \and
 \email{suravipatnaik1992@gmail.com}
 \and
 University of Otago, Dunedin, New Zealand \\ \email{stephen.cranefield@otago.ac.nz}}
\maketitle              
\begin{abstract}
 Researchers have long been interested in the role that norms can play in governing agent actions in multi-agent systems. Much work has been done on formalising normative concepts from human society and adapting them for the government of open software systems, and on the simulation of normative processes in human and artificial societies. However, there has been comparatively little work on applying normative MAS mechanisms to understanding the norms in human society.

 This work investigates this issue in the context of international politics. Using the GDELT dataset, containing machine-encoded records of international events extracted from news reports, we extracted bilateral sequences of inter-country events and applied a Bayesian norm mining mechanism to identify norms that best explained the observed behaviour. A statistical evaluation showed that the normative model fitted the data significantly better than a probabilistic discrete event model.

\end{abstract}
\section{Introduction} \label{sec:intro}

Norms are patterns of expected behaviour in human societies. In multi-agent systems, norms have been an active area of research where they have been used to facilitate desired agent-behaviour by constraining choices through prohibitions, obligations and permissions. Norms have been shown to facilitate social order \cite{conte2001} and also improve cooperation and coordination among agents \cite{shoham1997}, and an active research community has investigated many theoretical and practical aspects of normative reasoning in multi-agent systems~\cite{DBLP:conf/dagstuhl/2013dfu4}.

An emergent line of work is on how individual agents can use observations of agent societies to identify the norms that are prevalent in their society. Most work on observation-based norm learning has been limited to simulated datasets and simulation-based studies. This work aims to bridge the gap by inferring norms from a real dataset in the domain of international politics.

The Global Database of Events, Language and Tone (GDELT) \cite{Leetaru13gdelt:global} is a continuously updated geopolitical event database with over half a billion records. It is free and open source and the current version, GDELT 2.0, is updated every 15 minutes. The database includes an events table containing machine-coded data comprising 60 attributes for each event extracted from news reports (e.g.~the event type and countries involved)~\cite{GDELTCodeBookV2}, and has been used for studies such as predicting future violence levels in Afghanistan \cite{yonamine2013b}, and detecting protest events in the world \cite{Qiao2015}.

The objective of our research is to identify norms in international politics from the GDELT database. We will focus on the discovery of norms that can be viewed as latent variables \cite{Blei2014}. Currently, we do not aim to build a model to predict future behaviour in international politics, instead we focus on adapting a norm-learning technique that can extract norms from this database.

We consider the following research question: \textit{Can the GDELT database be better explained by a model combining probabilities and norms than by a purely probabilistic model?} Our research comprised five stages: i) collecting and pre-processing data from the GDELT database; ii) fitting a probabilistic model to serve as the baseline model; iii) defining a model to calculate the likelihood of observed sequences of events, given an assumed norm; iv) using this model to mine norms from the GDELT database based on Bayesian learning; and v) comparing the fit of the two models with the data.

The rest of the paper is outlined as follows. In \Cref{sec:priorwork} we discuss prior work on norm learning. \Cref{sec:dpp} discusses how we processed the GEDLT data to extract bilateral sequences of mutually relevant events. \Cref{sec:prob_model} presents the probabilistic model used as a baseline for our work, and \Cref{sec:norm_model} discusses the Bayesian approach to norm learning, our model for norms, and our model for normative reasoning in the form of a likelihood function for observations given a candidate norm. \Cref{sec:results,sec:evaluation} present the results of our experiments and a statistical evaluation of the increased explanatory power that our model gives compared to the baseline probabilistic model. Finally, we present our conclusions and suggestions for future work.

\section{Prior Work on Norm Learning}\label{sec:priorwork}

Norm learning (also known as norm identification or recognition) is an active area of research in multi-agent systems. Researchers have employed various techniques for inferring norms including association rule mining \cite{Savarimuthu2010,savarimuthu2013a}, case-based reasoning \cite{Balke2009,Campos2010}, reinforcement learning \cite{DBLP:conf/ijcai/SenA07}, inductive logic programming \cite{DomenicoEtAl_2011,DBLP:conf/aaai/TanBS19} and Bayesian inference \cite{DBLP:conf/ecai/CranefieldMOS16}. Most work in this area is limited to inferring norms from simulated agent societies. For example, the work of Savarimuthu et al.~\cite{Savarimuthu2010,savarimuthu2013a} inferred prohibition and obligation norms in simulated park littering and restaurant scenarios, while Sen and Airiau \cite{DBLP:conf/ijcai/SenA07}
investigated social learning by simulating repeated encounters in an abstract model of a traffic intersection. While there is value in simulation studies that demonstrate the feasibility of an approach and allow parameter tuning, the approaches become more beneficial when they are applied to real-world data to demonstrate their utility.

There has been some work on applying multi-agent systems norm-mining techniques to infer norms from real-world datasets. Researchers have explored the identification of software development norms from open source software repositories \cite{Dam2015}. Also, researchers have investigated a corpus of business contracts \cite{Gao2014} that contain explicit specifications of prohibitions and obligations. Norms in these domains are explicitly stated (e.g.~``One must not do X''). These approaches use features in sentences (e.g.~deontic modalities in the sentences such as \emph{must} and \emph{must not}) to infer norms. These works typically either use smaller datasets (e.g.~868 sentences in the work of Gao and Singh \cite{Gao2014}) and/or the information about a norm is found in a single document \cite{Avery2016}. In contrast, this paper deals with a very large dataset where norms lie hidden and norm-related information pertaining to a norm (e.g.~details about norm triggers, violations and sanctions) cannot be found in a single document, but is spread across different articles, potentially spanning a long period of time.

\section{Data Pre-processing}\label{sec:dpp}


The GDELT 2.0 \emph{events table} records events between pairs of actors. Each event has 34 attributes; we focus on GlobalEventID, Actor{\it N}CountryCode, Actor{\it N}Type1Code, Day, and EventCode (for $N\in\{1,2\}).$~\cite{GDELTCodeBookV2}. Event and actor types are encoded using the Conflict and Mediation Event Observations (CAMEO) coding scheme \cite{Schrodt2012}. CAMEO event codes have a hierarchical structure, with 20 ``root codes'' (in GDELT terminology) that are further specialised into ``base codes'', with some subdivided into a further level of detail. These are shown in~\Cref{table:root_codes}. For example, ``Albania on Friday denounced as an ugly crime Yugoslavia's suppression of ethnic Albanian unrest'' \cite{Schrodt2012} is encoded with base code 111: Criticize or Denounce, a subcode of root code 11: Disapprove. Its attributes record the two actors' countries and their types: both \emph{government}.

We considered events occurring from June 19, 2018 to June 20, 2019, recorded in 35039 data files, and retained only those involving two country actors (as opposed to other international organisations) with a primary role code of ``government''. We restricted our analysis to bilateral inter-country interactions, and used the CAMEO root codes only.

Our data pre-processing phase involved extracting these interactions and inferring which events are related to create sequences of related events. To do so we introduced the notion of \emph{mutual relevance} and a \emph{co-mention} relation. We defined two events to be co-mentioned if they appear in the same news source as recorded in the GDELT \emph{mentions table} \cite{GDELTCodeBookV2}. This relation on the set of events is denoted by $\sim_{cm}$. This relation is symmetric and reflexive by definition. We define the mutual relevance relation $\sim_{r}$ as the transitive closure of $\sim_{cm}$. The mutual relevance relation is an equivalence relation and thus partitions the set of events. As we are interested in bilateral events, we divide each partition into sub-partitions containing events involving a distinct pair of country actors. These sub-partitions, when ordered by date, form sequences of mutually relevant bilateral events. When two events occur on the same day we order them randomly. Each event in a sequence is represented by its CAMEO event code and actor 1 and actor 2 country names.

\begin{table}[t]
 \caption{CAMEO root event codes}
 \label{table:root_codes}
 \footnotesize
 \rule{\linewidth}{0.8pt}
 \begin{minipage}[t]{0.49\linewidth}
  \begin{tabularx}{\linewidth}[t]{rX}
   01 & Make public statement            \\
   02 & Appeal                           \\
   03 & Express intent to cooperate      \\
   04 & Consult                          \\
   05 & Engage in diplomatic cooperation \\
   06 & Engage in material cooperation   \\
   07 & Provide aid                      \\
   08 & Yield                            \\
   09 & Investigate                      \\
   10 & Demand                           \\
  \end{tabularx}
 \end{minipage}
 \begin{minipage}[t]{0.49\linewidth}
  \begin{tabularx}{\linewidth}[t]{rX}
   11 & Disapprove                             \\
   12 & Reject                                 \\
   13 & Threaten                               \\
   14 & Protest                                \\
   15 & Exhibit military posture               \\
   16 & Reduce relations                       \\
   17 & Coerce                                 \\
   18 & Assault                                \\
   19 & Fight                                  \\
   20 & Engage in unconventional mass violence \\
  \end{tabularx}
 \end{minipage}
 \rule{\linewidth}{0.8pt}
\end{table}

Our initial generation of sequences resulted in some that seemed overly long\footnote{Some sequences contained as many as 20,000 events.} to comprise only mutually relevant events. We found two issues contributing to this problem. Many BBC news reports had non-unique mention identifiers.
We therefore filtered out events with ``BBC'' in their mention identifiers before determining the mutually relevant event sets. We also found that some events were highly mentioned, possibly as the general background to news stories. These functioned as hubs linking many events together via the co-mention relationship. We resolved this problem by `cloning' events with more than 250 mentions, i.e.~for each mention of a highly mentioned event we generated a separate copy of that event with a unique mention identifier. After these two steps, our processed data consisted of 513,906 sequences with the maximum sequence length of 161.

\section{Baseline probabilistic model}\label{sec:prob_model}

We take a Bayesian approach to norm mining, which we explain in detail in \Cref{sec:norm_model}. This requires computing the likelihood of an observed sequence of events, given each norm hypothesis (including the hypothesis that there are no norms). Prior work on Bayesian learning of norms assumed knowledge of the plans that govern agents' public behaviour \cite{DBLP:conf/ecai/CranefieldMOS16}---this provided a model of the possible agent behaviours in the absence of normative reasoning. In this work, as we do not have a plan-based model of international political interactions, we fit a probabilistic model to the set of event sequences resulting from our data pre-processing. This is used when defining the likelihood of observed sequences given norm hypotheses, and also serves as a baseline model for the evaluation of our normative model, as described in \Cref{sec:evaluation}. We chose to use the libPLUMP implementation\footnote{\url{http://www.gatsby.ucl.ac.uk/~ucabjga/libplump.html}} of the \emph{sequence memoizer} (SM) \cite{SequenceMemoizerCACM}, due to Murphy's description of the SM as ``the best-performing language model'' \cite[p.595]{Murphy:2012:MLP:2380985}. The SM is a Bayesian non-parametric model that can be trained to learn a conditional distribution for the next symbol given all previous symbols~\cite{DBLP:conf/nips/GasthausT10}, i.e.~a set of conditional probability distributions of the form $p(x_{i+1}|x_1,\cdots,x_i)$. Here, $x_1, \cdots, x_i$ is a sequence of observed symbols (the \emph{context}), and the distribution tells us the probability of symbol $x_{i+1}$ being the next symbol to appear after the given context. In the SM, the length of the context sequences is not bounded.



The event sequences extracted from the GDELT dataset contain events represented as triples $\langle rc, c_1, c_2 \rangle$, where $rc$ is a CAMEO root code and $c_1$ and $c_2$ are the country identifiers appearing as the event's Actor1CountryCode and Actor2CountryCode fields. As we are attempting to learn generic norms that apply to all countries, we do not retain the country identifiers when training the SM. However, we must preserve the directionality of the events between the two countries: there is an important difference in the relative directions of the events in the first sequence below compared to those in the second and third sequences.
\[ \langle rc_1, c_1, c_2 \rangle, \langle rc_2, c_1, c_2 \rangle \quad
 \langle rc_1, c_1, c_2 \rangle, \langle rc_2, c_2, c_1 \rangle \quad
 \langle rc_1, c_2, c_1 \rangle, \langle rc_2, c_1, c_2 \rangle
\]
We therefore write each event as a combination of an event root code and a \emph{direction}, where by convention the first event in a sequence is taken to be in the ``forwards'' direction. We denote a directed event by a pair $\langle \mathit{dir},\mathit{code} \rangle$, giving the following representation for the second sequence above: $\langle \mathit{F},\mathit{rc_1} \rangle, \langle \mathit{B},\mathit{rc_2} \rangle$ where $F$ denotes ``forwards'' and $B$ denotes ``backwards''.
For simplicity, we will denote a directed event by single variable, e.g.~$e$, where we do not need to make the direction explicit.

While training the SM, we generate a second instance of each sequence in which the event directions are reversed. This is because we intend the trained SM to be used to look up likelihoods for observed event sequences (which will be encoded in the same way as the training data). If the SM does not have a stored context that completely matches the observed sequence, it may need to find the longest observation suffix that it has stored in its internal context tree. If all SM contexts began with events in the forward direction (our convention during the original sequence generation), then this suffix matching would not work for observation suffixes that have an odd number of initial events omitted. Finally, we append a special end symbol to each sequence. 

\section{Normative Model}\label{sec:norm_model}

We follow the Bayesian approach to norm learning \cite{DBLP:conf/ecai/CranefieldMOS16}. Given a set of norm hypotheses, for every observed event sequence $\sigma$ in our dataset and each norm hypothesis $h$ we calculate the likelihood of the observation given the hypothesis: $p(\sigma|h)$. Bayes' Rule (the odds form of Bayes' Theorem) is then used to update the odds of two hypotheses given the observation:
\begin{align}\label{eq:bayes}
 O(h_1{:}\,h_2 | \sigma) = O(h_1{:}\,h_2) \frac{p(\sigma|h_1)}{p(\sigma|h_2)}
\end{align}
Here, we multiply the prior odds $O(h_1{:}\,h_2)$ by the ratio of the likelihoods of the observed sequence under the two hypotheses, to give the posterior odds \footnote{The prior odds are initialised to 1 for all of the norms we consider. However, this choice has no significant effect as we are interested in the \emph{relative} posterior odds.}.

We reason with odds because, in general, candidate norms do not form a set of mutually exclusive and collectively exhaustive hypotheses, which is necessary to normalise probabilities when using Bayes' Theorem \cite{DBLP:conf/ecai/CranefieldMOS16}. In particular, we calculate the odds of our norm hypotheses relative to the null hypothesis that there are no norms ($h_0$), and we work in log space. After initialising the prior log odds for all norm hypotheses to a uniform value, for each sequence $s$ we calculate the log likelihood of $s$ under all hypotheses. We then update the log odds for each hypothesis $h$ by adding the difference between $\mathop{log} p(s|h)$ and $\mathop{log} p(s|h_0)$.\footnote{Note that this approach only reasons about single-norm hypotheses. Based on the log odds of individual norms, it would be possible to select combinations of most likely norms to form multi-norm hypotheses, but this requires a more complex model of observation likelihood than we have at present.}

\subsection{Norm hypotheses}
\label{sec:norms}

\newcommand{\ec}{\mathit{ec}}
\newcommand{\cec}{\mathit{cec}}
\newcommand{\parity}{\mathit{rel\_dir}}
\newcommand{\same}{+}
\newcommand{\different}{-}
\newcommand{\defeq}{\stackrel{\mathclap{\text{def}}}{=}}
\newcommand{\code}{\operatorname{code}}
\newcommand{\dir}{\operatorname{dir}}
\newcommand{\F}{\operatorname{F}}
\newcommand{\B}{\operatorname{B}}

\newcommand{\nnext}{\LTLcircle}
\newcommand{\prev}{\LTLcircleminus}
\newcommand{\eventually}{\LTLdiamond}
\newcommand{\always}{\LTLsquare}

We use the following language to define our norm hypotheses space:
\begin{itemize}
 \item
       $O(\ec)$: an unconditional obligation to perform an event with event code $\ec$.
 \item
       $O(\cec, \ec, \parity)$: a conditional obligation to perform an event with event code $\ec$ if a prior condition event with event code $\cec$ has occurred, and the two events have the relative direction specified by $\parity$ (either `same' or `different', denoted $\same$ and $\different$, respectively).
 \item
       $P(\ec)$: This represents an unconditional prohibition of events with event code $\ec$.
 \item
       $P(\cec, \ec, \parity)$: a conditional prohibition of events with event code $\ec$ if a prior condition event with event code $\cec$ has occurred, and the two events have the relative direction specified by $\parity$.
\end{itemize}

For conditional norms, the relative direction constraints are necessary to specify which country is subject to the norm once the condition occurs: the country that performed the event triggering the norm, or the other one.

The semantics of these norm types are defined using linear temporal logic interpreted over event sequences. For example:
\begin{align*}
 O(\cec, \ec, \same) \:\defeq
  & {}\quad \always\: (\code(\cec) \land \dir(\F) \rightarrow \nnext\eventually\, (\code(\ec) \land \dir(\F))) \land {} \\
  & {}\quad \always\: (\code(\cec) \land \dir(\B) \rightarrow \nnext\eventually\, (\code(\ec) \land \dir(\B)))
\end{align*}
where $\code$ and $\dir$ are predicates recording the event code and direction of an event, and $\F$ and $\B$ denote the forwards and backwards event directions. $O(\cec, \ec, \different)$ and the other norm types are defined in a similar fashion.

For our experiments, our norm hypothesis set was formed by instantiating the $O(\cec, \ec, \parity)$, $O(\ec)$, $P(\ec)$ and $P(\cec, \ec, \parity)$ norm types with $\cec$ and $\ec$ ranging over the 20 CAMEO root codes and $\parity \in \{\same, \different\}$. We therefore had 1640 norm hypotheses.

\subsection{Computing observation likelihood ratios}

Consider \Cref{eq:bayes}. Let $\bm{\sigma} = \left( \sigma_1, \sigma_2, \dotsc, \sigma_{N} \right) $ be our dataset with $N=513,906$ event sequences. Then, as each sequence is independently observed, we can express the conditional probability of the data given a norm hypothesis $h$ as follows:
\begin{align}\label{eq:prod}
 p(\bm{\sigma}|h) = \prod_{i=1}^{N} p(\sigma_i|h)
\end{align}
Applying \Cref{eq:bayes}, taking logarithms, and considering the odds of a norm hypothesis holding versus the no-norm null hypothesis $h_0$, we calculate the log odds of the complete dataset as the sum of the prior log odds and the \emph{log likelihood ratios} of each sequence under the two hypotheses.
\begin{align}\label{eq:log_odds}
 log(O(h{:}\,h_0 | \bm{\sigma})) & = log(O(h{:}\,h_0)) +\sum_{i=1}^{N} \left(log(p(\sigma_i|h)) - log(p(\sigma_i|h_0))\right)
\end{align}

\subsection{A state machine for norms}


To calculate the log likelihood of an observed sequence given a norm, we need a model of how the countries act in the presence of norms. This has two parts. In this subsection we present a state machine that models the changes of state of a norm as a sequence of events are observed. In the next subsection we define the likelihood of a sequence of events given a norm hypothesis, which is defined recursively while tracking the norm state.

\begin{figure}[th]
 \centering
 \includegraphics[height=0.4\textheight]{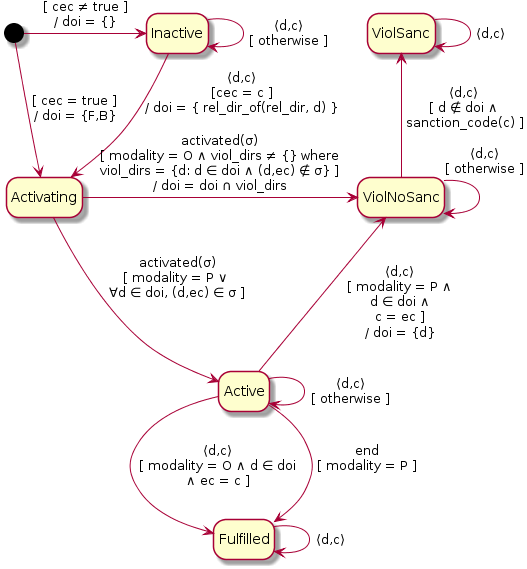}
 \caption{The norm state machine}
 \label{fig:nstate}
\end{figure}

The norm state machine is shown in UML notation in \Cref{fig:nstate}\footnote{The transition annotations have the format $\mathit{trigger} [\mathit{guard}] / \mathit{action}$. Transitions from the initial state have no trigger, and guards and actions are optional.}. A separate state machine is created for each observed sequence and norm hypothesis. The variable \verb|modality| is instantiated with the value \verb|O| or \verb|P|, representing an obligation or prohibition, respectively. The variables \verb|cec|, \verb|ec| and \verb|rel_dir| record the corresponding norm parameters from \Cref{sec:norms}. For unconditional norms, \verb|cec| is set to \verb|true|. The observed events are passed to the state machine as a $\langle \mathit{direction}, \mathit{event\_code}\rangle$ pair, followed by an \verb|end| event. The state machine uses the variable \verb|doi| to track the ``directions of interest'' for checking events for norm fulfilment, violation and sanctioning\footnote{To simplify the model, only the first instantiation, fulfilment, violation and sanction of a norm within a sequence are tracked by the state machine.}. For unconditional norms, \verb|doi| is initially the set $\{\F,\B\}$ (i.e., both forward and backwards), while for conditional norms, a singleton direction set is assigned to \verb|doi| based on the direction of the triggering event, and the norm's $\parity$.

We categorise the following CAMEO root event codes as sanction events, as they signal disapproval towards another country: 11 (Disapprove), 12 (Reject) and 16 (Reduce relations)\footnote{We omit conflict event types (such as Threaten, Assault and Fight) from this list, as we consider these to be events that are likely to be subject to norms, rather than reactions to norm violations.}.

We considered it impractical to include deadlines as parameters of our obligation norm hypotheses as this would greatly enlarge the norm hypothesis space and require us to include time stamps in events.
Thus, a transition from \verb|Activating| to \verb|ViolNoSanc| is made immediately if an obligation is violated by the remaining events in the sequence. In this case, sanctions will be recognised however soon they appear after norm activation.

\subsection{Observation likelihood} \label{sec:likelihood}
In this section, we define the likelihood of an observed sequence of events, given a candidate norm. This encodes our model of how norms influence the behaviour of interacting countries by varying the background probability of events (as modelled by the sequence memoizer).

Given a sequence of events $\sigma$ (including the $\mathrm{end}$, symbol added before SM training), and the null hypothesis ($h_0$) that there is no norm, we define:
\begin{equation}\label{eq:top_level1}
 p(\sigma|h_0) = \prod_{i=1,\dots,|\sigma|} p_{SM}(\sigma_i|\sigma^{\leq i})
\end{equation}
\noindent where $p_{SM}(e|h)$ denotes the probability of the event $e$ given context history $h$ returned from a sequence memoizer trained on the entire dataset, $\sigma_i$ denotes the $i^{\mathrm{th}}$ element of $\sigma$, $\sigma^{\leq i}$ is the prefix of $\sigma$ of length $i$, and $|\sigma|$ is the length of $\sigma$.

\newcommand{\nsm}{\mathit{nsm}}
\newcommand{\modality}{\mathit{modality}}
\newcommand{\diroi}{\mathit{doi}}
\newcommand{\nstate}{\mathit{state}}
\newcommand{\append}{{:}}
\newcommand{\req}[2]{\exists \langle d,e \rangle \in #1\times#2}
\newcommand{\avoid}[2]{\nexists \langle d,e \rangle \in #1\times#2}
\newcommand{\hd}{\mathop{hd}}
\newcommand{\tl}{\mathop{tl}}

For other (non-null) norm hypotheses, $n$, we define:
\begin{equation} \label{eq:top_level2}
 p(\sigma|n) = p(\sigma|\nsm(n),\epsilon)
\end{equation}
where $\nsm(n)$ is a new instance of the norm state machine for norm $n$, and $\epsilon$ is the empty sequence, representing the sequence history prior to $\sigma$. In the following, we define the conditional probability $p(\sigma|s,h)$ appearing on the right hand side of Equation~\ref{eq:top_level2} as a recursive function, with separate cases for the various states of the norm state machine\footnote{We perform the calculations in log space, resulting in a log likelihood, but for simplicity of presentation we do not show this.}. Given a norm state machine $s$, we write $s.\modality$, $s.\ec$, $s.\nstate$, and $s.\diroi$  to refer to the state machine's norm modality, norm event code, the current state and the value of the \verb|doi| variable.
We write $\hd(\sigma)$ and $\tl(\sigma)$ to denote the head and tail of the sequence $\sigma$, and $\sigma\append{}e$ for $\sigma$ with event $e$ appended. Given a sequence $\sigma$ and state machine $s$, we write $s^\prime$ as an abbreviation for $s.\mathit{receive}(\hd(\sigma))$, i.e.~the state machine resulting from sending $\hd(\sigma)$ to $s$. In other words, wherever $s^\prime$ appears on the right hand side of an equation, the norm state machine has invoked to update the state of the norm based on $\hd(\sigma)$. For case $s.\nstate = \text{\mdseries\tt Activating}$, we define $s^a = s.\mathit{receive}(\hd(\mathrm{activated}(\sigma)))$.



\newcommand{\compprob}{{\mathit p\_comp}}
\newcommand{\sancprob}{{\mathit p\_sanc}}

\begin{description}[leftmargin=0pt]

 \item[Case $\sigma=\langle\mathrm{end}\rangle$:]
       \begin{equation*} \label{eq:end-case}
        p(\langle\mathrm{end}\rangle|s,h) = p_{SM}(\mathrm{end}| h)
       \end{equation*}

 \item[Case $s.\nstate \in \{\text{\mdseries\tt Inactive},\text{\mdseries\tt Fulfilled},\text{\mdseries\tt ViolSanc}\}$:]
       \begin{equation*} \label{eq:default-case}
        p(\sigma | s, h) = p_{SM}(\hd(\sigma) | h) \; p(\tl(\sigma) | s^\prime, h\append\hd(\sigma))
       \end{equation*}

 \item[Case $s.\nstate = \text{\mdseries\tt Activating}$:]
       In this case we evaluate the conditional probability under three mutually exclusive additional assumptions: that the agent will be compliant with the norm, non-compliant but unsanctioned, and non-compliant and sanctioned. Two of these three assumptions will turn out be inconsistent with $\sigma$ and the norm embodied by the state machine $s$, and the corresponding conditional probabilities will return $0$. We write ``comp'' and ``sanc'' to denote the events of compliance and sanctioning occurring.
       \begin{align*} \label{eq:activating}
        \begin{split}
         p(\sigma | s, h) = {} & \compprob(n)^{|s.\diroi|} \, p(\sigma | s^a, h, \mathrm{comp})
         + {} \\
         & (1-\compprob(n)^{|s.\diroi|}) \, \sancprob(n) \, p(\sigma | s^a, h, \lnot \mathrm{comp}, \mathrm{sanc}) + {} \\
         & (1-\compprob(n)^{|s.\diroi|}) \, (1-\sancprob(n)) \, p(\sigma | s^a, h, \lnot \mathrm{comp}, \lnot \mathrm{sanc})
        \end{split}
       \end{align*}

       Here, we assume that there are two key decisions that a country must make within a bilateral interaction in the presence of a norm: whether to \emph{comply} with the norm, and whether to \emph{sanction} a violation of the norm by the other party.
       The \emph{comply} and \emph{sanction} decisions are governed by probabilities $\compprob(n)$ and $\sancprob(n)$, which we learn from the data on a per-norm basis. We count how many times each norm hypothesis is triggered\footnote{In the case of unconditional norms, this is once for every observation.}, violated, fulfilled and sanctioned across all the sequences in the dataset, and use add-one smoothing to address any zero counts.
       Then, for each norm hypothesis $h$, we calculate the following empirical probabilities of compliance and of sanctioning for norm violations. These are used when computing the log likelihood of observed event sequences given norm hypotheses.
       \newcommand{\triggerings}{{\mathit t}}
       \newcommand{\violations}{{\mathit v}}
       \newcommand{\sanctions}{{\mathit s}}
       \newcommand{\issanction}{\operatorname{is\_sanction}}
       \begin{gather*}
        \compprob(n) = \frac{\#\triggerings(n) - \#\violations(n) + 1}{\triggerings(n) + 2} \quad
        \sancprob(n) = \frac{\#\sanctions(n) + 1}{\#\violations(n) + 2}
       \end{gather*}
       where $n$ is a norm hypothesis, $\mathit t$, $\mathit v$, $\mathit s$ denote triggerings, violations and sanctions, and $\#$ abbreviates ``number of''.

       In the case of unconditional norms, a sequence in which compliance is observed means that both agents have decided to be compliant. Thus, in the equation for $p(\sigma | s, h)$ 
       we raise $\compprob$ to the power of $|s.\diroi|$ (the number of ``directions of interest'', which is 2 in the case of unconditional norms).


 \item[Case $s.\nstate = \text{\mdseries\tt Active}$:]
       \begin{align*}
        \begin{split}
         p(\sigma | s, h, \mathrm{comp})      & =
         \begin{cases}
          p_1 \, p_3 & \text{if $s.\modality = \mathrm{O}$ and $\exists d\in s.\diroi,\, \langle s, s.\ec \rangle \in \sigma$}     \\
          p_2 \, p_3 & \text{if $s.\modality = \mathrm{P}$ and $\forall d\in s.\diroi,\, \langle s, s.\ec \rangle \not\in \sigma$} \\
          0          & \text{otherwise}
         \end{cases}
        \end{split}
       \end{align*}
       \noindent and
       \begin{align*}
        \begin{split}
         p(\sigma | s, h, \lnot \mathrm{comp},v) & =
         \begin{cases}
          p_1\, p_4 & \text{if $s.\modality = \mathrm{P}$ and $\exists d\in s.\diroi,\, \langle s, s.\ec \rangle \in \sigma$}     \\
          p_2\, p_4 & \text{if $s.\modality = \mathrm{O}$ and $\forall d\in s.\diroi,\, \langle s, s.\ec \rangle \not\in \sigma$} \\
          0         & \text{otherwise}
         \end{cases}
        \end{split}
       \end{align*}
       \noindent where $v$ can be either $\mathrm{sanc}$ or $\neg \mathrm{sanc}$ and:
       \begin{align*}
        p_1 & \equiv p_{\mathit{incl}(s.\diroi,\mathit{\{s.\ec\}})}(\hd(\sigma)|h), \;
        p_2 \equiv p_{\mathit{excl}(s.\diroi,\mathit{\{s.\ec\}})}(\hd(\sigma)|h)       \\
        p_3 & \equiv p(\tl(\sigma) | s^\prime, h\append\hd(\sigma), \mathrm{comp}), \;
        p_4 \equiv p(\tl(\sigma) | s^\prime, h\append\hd(\sigma), \lnot \mathrm{comp},v)
       \end{align*}

       \noindent $p_1$ and $p_2$ are defined in terms of some specialised conditional probability distributions:
       \[
        p_{\mathit{incl}(\diroi,\mathit{codes})}(e|h) \equiv {}
        p(e\,|\,h,\, h\append e \sim \{\sigma\in D: \exists \langle d,c \rangle\in \diroi\times\mathit{codes}, \langle d,c \rangle\in \sigma\})
       \]
       \[
        p_{\mathit{excl}(\diroi,\mathit{codes})}(e|h) \equiv {}
        p(e\,|\,h,\, h\append e \sim \{\sigma\in D: \nexists \langle d,c \rangle\in \diroi\times\mathit{codes}, \langle d,c \rangle\in \sigma\})
       \]

       $p_{\mathit{incl}(\diroi,\mathit{codes})}(e|h)$ is the probability of the directed event $e$ occurring after event history $h$ according to a language model trained on sequences from our dataset ($D$) that \emph{include} an event with direction in $\diroi$ and an event code in $\mathit{codes}$. This probability is used to determine the probability of an event under the assumption that there is an obligation norm and the norm is known to be active, or that a norm has been violated and will be sanctioned. In these cases, we know that the obligated event or a sanction should be one of the remaining events in the sequence, i.e.~sequences that do not include such an event have a probability of zero. Therefore, to ensure we are using a correctly normalised probability distribution we must use a sequence memoizer trained only on data that excludes these sequences.

       We are interested in $p_{\mathit{incl}(\diroi,\mathit{codes})}(e|h)$ for cases where $\mathit{codes}$ is a singleton set or the set of all sanction event codes. To implement $p_{\mathit{incl}(\cdots)}$, we train sequence memoizers for each case, using only sequences containing a forward-directed event with an event code in the $\mathit{codes}$ set for that case. Given these, we can compute this probability for backwards-directed events $de$ by reversing the event directions in both $de$ and $h$\footnote{Recall from \Cref{sec:prob_model} that we train sequence memoizers with two copies of each event sequences: one with its default directions (where the first event is taken to be ``forwards'' and a direction-reversed copy of the sequence).}.

       Note that in cases of conditional norms, this probability is used as an approximation. We would actually need to train the sequence memoizer on sequences that exclude such events only \emph{after} the norm is activated by a norm's specific triggering event. This would involve using many more sequence memoizers, one for each pair of event codes (a triggering event and an obligated event), so we have chosen to approximate it. Note that the proposed sequence memoizer \emph{underestimates} the probability as compared to that which trains on every pair of event codes, since it is trained on a superset of sequences. This lower probability dampens the log odds of the norm, and thus our estimates of the log odds of conditional obligations are conservative.

       $p_{\mathit{excl}(\diroi,\mathit{codes})}(e|h)$ is the probability of the directed event $e$ occurring after event history $h$, according to a language model trained only on sequences from our dataset ($D$) that \emph{exclude} events with direction in $\diroi$ and an event code in $\mathit{codes}$. This probability is used to determine the probability of an event under the assumption that there is a prohibition norm and the norm is known to be active, or that a norm has been violated but will not be sanctioned. This probability distribution is implemented by modifying the probabilities returned by the default sequence memoizer $p_{SM}$ (which is trained on the full set of sequences):
       \begin{align*}
        p_{\mathit{excl}(\diroi,\mathit{codes})}(\langle d,c \rangle|h) =
        \begin{cases}
         \; 0                                                                                               & \begin{minipage}[c]{0pt}\begin{tabbing}if $d\in\diroi \land{}$ \\$c\in\mathit{codes}$\end{tabbing}\end{minipage} \\
         \; p_{SM}(\langle d,c \rangle | h)\: /                                                                                          \\
         \quad (1 - \sum_{d^\prime\in\diroi, c^\prime\in\mathit{codes}}\, p_{SM}(\langle d',c' \rangle)|h)) & \text{otherwise}
        \end{cases}
       \end{align*}

 \item[Case $s.\nstate = \text{\mdseries\tt ViolNoSanc}$:]
       \begin{align*}
        \begin{split}
         p(\sigma | s, h, \mathrm{comp})           & =
         0 \quad                            \text{(\texttt{ViolNoSanc} inconsistent with `comp' assumption)} \\
         p(\sigma | s, h, \lnot \mathrm{comp}, \mathrm{sanc})   & =
         \begin{cases}
          0          & \text{if $\hd(\sigma)$ is not a sanction in a direction in $s.\diroi$} \\
          p_1 \, p_3 & \text{otherwise}
         \end{cases}
        \end{split}
       \end{align*}
       \noindent and
       \begin{align*}
        \begin{split}
         p(\sigma | s, h, \lnot \mathrm{comp}, \neg \mathrm{sanc}) & =
         \begin{cases}
          0          & \text{if $\hd(\sigma)$ is a sanction in a direction in $s.\diroi$} \\
          p_2 \, p_4 & \text{otherwise}
         \end{cases}
        \end{split}
       \end{align*}
       \noindent where
       \begin{align*}
        p_1 & \equiv p_{\mathit{incl}(s.\diroi,\mathit{sanction\_codes})}(\hd(\sigma)|h),\;
        p_2 \equiv p_{\mathit{excl}(s.\diroi,\mathit{sanction\_codes})}(\hd(\sigma)|h)                     \\
        p_3 & \equiv p(\tl(\sigma) | s^\prime, h\append\hd(\sigma), \lnot \mathrm{comp}, \mathrm{sanc}),\;
        p_4 \equiv p(\tl(\sigma) | s^\prime, h\append\hd(\sigma), \lnot \mathrm{comp}, \lnot \mathrm{sanc})
       \end{align*}
\end{description}

\begin{table*}[t]
 \caption{Top six norms}
 \label{table:top-norms}
 \begin{tabularx}{\linewidth}{l *{8}{C}}
  \toprule
  {}                  & log odds & triggers & fulfilled & viols  & non sanc. viols & sanc. viols & pcomp & psanc \\
  \midrule
  $O(4)$              & 75355.1  & 513906   & 115633    & 398273 & 349043          & 4230.0      & 0.225 & 0.011 \\
  $O(4,4,\different)$ & 65428.0  & 232767   & 115633    & 117134 & 114921          & 2213.0      & 0.497 & 0.019 \\
  $O(3)$              & 21550.6  & 513906   & 16384     & 497522 & 495551          & 1971.0      & 0.032 & 0.004 \\
  $O(3,3,\different)$ & 14569.4  & 56951    & 16384     & 40567  & 39548           & 1019.0      & 0.288 & 0.025 \\
  $O(5,5,\different)$ & 13814.1  & 66766    & 12618     & 54148  & 52404           & 1744.0      & 0.189 & 0.032 \\
  $O(4,3,\same)$      & 12092.1  & 232767   & 4750      & 228017 & 226077          & 1940.0      & 0.020 & 0.009 \\
  \bottomrule
 \end{tabularx}
\end{table*}

\section{Results}\label{sec:results}

To learn norms from the data, we estimated the probability of compliance and sanctioning for each norm, as discussed in \Cref{sec:likelihood}, and ran the Bayesian inference procedure on our event sequence dataset and hypothesis set. This resulted in 173 norms with posterior log odds greater than 0 (i.e.~odds greater than 1), and which were therefore found to be more likely hypotheses than the null hypothesis (that there are no norms).
Among these 173 norms, there are 154 conditional obligation norms, 16 conditional prohibition norms and 3 unconditional obligation norms.
The top six norms are shown in \Cref{table:top-norms}, along with their posterior log odds, counts of triggerings, fulfilments, violations and sanctions, and the inferred probability of compliance and sanctioning.
These top norms can be understood as follows:

\begin{itemize}
 \item \textbf{$O(4)$ :} The CAMEO root code 4 as seen in \Cref{table:root_codes} stands for `Consult'. This norm is an unconditional obligation to consult the other country. This norm is complied with 22.5\% of the time which accounts for its high log odds. The violations in this norm were rarely sanctioned.
 \item \textbf{$O(4,4,\different)$ :} This is triggered by a consultation, and the other party is then obliged to consult. More than half the time it is violated, but some of the violations (1.8\%) are sanctioned. The 49.7\% fulfilment rate and the sanctions account for this norm's very high log odds.
 \item \textbf{$O(3)$, $O(3,3,\different)$, $O(5,5,\different)$ :} Much like the above two norms, these norms have high log odds. The first is the unconditional obligation to `Express Intent to Cooperate'. The second is triggered by an expression of intent to cooperate, obligates the other party to express intent to cooperate in return. The third norm is the obligation to engage in diplomatic cooperation once the other part engages in diplomatic cooperation.
 \item \textbf{$O(4,3,\same)$ :} This norm suggests that when a party consults the other party, it is also likely to express intent to cooperate.
\end{itemize}

\section{Evaluation}\label{sec:evaluation}

\newcommand{\LRT}{\mathit LRT}

\begin{figure}[tb]
 \centering
 \includegraphics[width=0.5\linewidth]{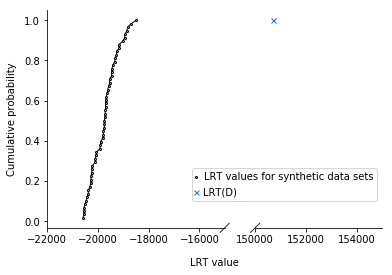}
 \caption{$\LRT(D)$ and the empirical distribution of $\LRT(d)$}
 \label{fig:p-value}
\end{figure}

To address our research question (\emph{Can the GDELT dataset be better explained by a model combining event probabilities and explicit norms than by a probabilistic model?}), we compare the total likelihood of the event sequence dataset under two models:
\begin{description}
 \item[$M_0$:] the probabilistic model embodied by the SM trained on all sequences, and
 \item[$M_1$:] a model that uses normative reasoning about the top norm to modify the SM's probabilities (as presented in \Cref{sec:likelihood}).
\end{description}

We denote the log likelihood of an event dataset $d$ under these two models as $L_0(d)$ and $L_1(d)$, respectively. We write $D$ to denote our GDELT-derived sequence dataset.

The total log likelihoods of all sequences in our GDELT-derived dataset under the two models were $L_1(D) - L_0(D) = 75355.051837$. Thus, $M_1$ is a better fit to the data than $M_0$. In order to determine whether the difference in log likelihood is statistically significant, we performed a likelihood ratio test. This test compared two hypotheses:
\begin{description}
 \item[$H_0$:] Model $M_0$ is the true model underlying the data. This is the null hypothesis.
 \item[$H_1$:] Model $M_1$ is the true model underlying the data.
\end{description}

Under certain conditions related to model nesting and smoothness of the likelihood function, the null hypothesis implies that the likelihood ratio test statistic $\LRT(d) = -2 (L_0(d) - L_1(d)) = 2 (L_1(d) - L_0(d))$ approximates a $\chi^2$ distribution for large sample sizes, with degrees of freedom equal to the difference in the number of free parameters between the two models~\cite{Wilks1938}.
A standard one-tailed significance test can then be used to reject $H_0$ if $LRT(D)$ is less than or equal to the $\chi^2$ value at the desired significance level. However, the conditions allowing the use of the $\chi^2$ test do not apply in our case, so we compute an empirical distribution of the $\LRT$ statistic.

We used the posterior distribution of sequences, i.e.~the trained SM, to generate $58$ synthetic datasets\footnote{The number was constrained by the highly time-intensive process of looking up the SM's conditional probabilities to generate and evaluate the likelihood of these large datasets.} with the same size as the real data set (i.e. 513,906 sequences) under the null hypothesis.\footnote{Murphy discusses the relationship between sampling from the posterior and the well known \emph{bootstrap} method for approximating the sampling distribution of a test statistic~\cite[p.192]{Murphy:2012:MLP:2380985}.}
Sequences were sampled until the end symbol was reached from the sequence memoizer's conditional predictive distributions: $p(e|\sigma)$, where $e$ ranged over all event root codes and $\sigma$ is a known prior sequence of events.

We calculated $LRT(d)$ for each synthetic dataset and the p-value for the observed value $LRT(D)$.
It was found that in each of these datasets, $LRT(d)$ was negative, indicating that the language model learned by the sequence memoizer is not sufficient to explain the top norm extracted from the real data.
The p-value is the probability that $\LRT(d) \geq \LRT(D), \, d \neq D$. \Cref{fig:p-value} shows the empirical distribution of $\LRT(d)$. As $\LRT(D)$ exceeds $\LRT(d)$ for all synthetic datasets $d$, the p-value is less than $\frac{1}{58} \approx 0.017$, meaning that we can reject the null hypothesis at the 0.1 significance level. Thus, we can answer our research question in the affirmative: our model including the top norm explains the GDELT dataset better than when using the sequence memoizer alone.


\section{Conclusions and Future Work}


We have presented a methodology for mining norms from a global political event database using a Bayesian learning approach, and an algorithm for computing the log likelihood of observed sequences of actions given a norm hypothesis and a background probabilistic model of event sequences.
A statistical evaluation showed that a model of event sequence likelihood that enhances the probabilistic model with normative reasoning fits the data significantly better than the baseline probabilistic model.

The primary contribution of this research is to propose a methodology for mining norms in international relations. With the growing presence of automated event datasets such as GDELT, research in this direction will provide tools for researchers and international organisations like the United Nations and the Inter-Parliamentary Union to gain a better understanding of the tacit norms that govern international relations. This work also provides a demonstration of how norm-learning techniques from multi-agent systems can apply to real-world human societies.

There are a number of limitations to our current approach that we plan to address in future work. We have focused on country actors with a government role, which could easily be extended to include other international organisations such as the United Nations, and police, military and intelligence agency roles.

We mined for norms using the full set of bilateral event sequences, so our inferred norms are those that apply globally, to all countries. It would be interesting to look for norms that are specific to particular groups of countries.

Our dataset was limited to a one-year period. A year is a short time in politics, and more interesting norms could be found from a dataset spanning a longer period, although this will require significantly greater computation time.

The detection of mutually relevant events relies on the ability of GDELT's automated event extraction techniques to assign the same identifier to an event across all mentions of it. The ability of GDELT's automated systems to uniquely identify events in this way has been questioned~\cite{UlfelderBlog2014}, and this warrants further investigation.



A distinction has been made in the literature between ``normal'' and ``normative'' behaviour \cite[Chap.~1]{Kelsen1990}. It is argued that for observed ``normal'' behaviour to be considered normative, it must have an explicit internal representation as a norm (rather than simply being copied behaviour), and be the subject of social processes that account for its emergence. The external view of international events provided by event databases such as GDELT does not allow assessment of how explicitly the actors may be aware of the norms, and the only visible social mechanism is sanctioning. As sanctions were present for all of our inferred norms, there is reason to consider that these `norms' are truly normative rather than simply normal behaviour. However, we therefore see our research as only a first step in an investigation that will require input from social scientists to validate and refine our findings.

%
%
%
\bibliographystyle{splncs04}
\bibliography{gdelt-norm-mining,bibliography}

\begin{thebibliography}{10}
\providecommand{\url}[1]{\texttt{#1}}
\providecommand{\urlprefix}{URL }
\providecommand{\doi}[1]{https://doi.org/#1}

\bibitem{DBLP:conf/dagstuhl/2013dfu4}
Andrighetto, G., Governatori, G., Noriega, P., van~der Torre, L.W.N. (eds.):
  Normative Multi-Agent Systems, Dagstuhl Follow-Ups, vol.~4. Schloss Dagstuhl
  - Leibniz-Zentrum fuer Informatik (2013)

\bibitem{Avery2016}
Avery, D., Dam, H.K., Savarimuthu, B.T.R., Ghose, A.: Externalization of
  software behavior by the mining of norms. In: 13th International Conference
  on Mining Software Repositories. pp. 223--234. ACM (2016)

\bibitem{Balke2009}
Balke, T., Novais, P., Andrade, F.C.P., Eymann, T.: From real-world regulations
  to concrete norms for software agents: a case-based reasoning approach. In:
  International Workshop on Legal and Negotiation Decision Support Systems
  (LDSS). pp. 14--27. CEUR Workshop Proceedings (2009)

\bibitem{Blei2014}
Blei, D.M.: Build, compute, critique, repeat: Data analysis with latent
  variable models. Annual Review of Statistics and its Application
  \textbf{1}(1),  203--232 (2014)

\bibitem{Campos2010}
Campos, J., L{\'o}pez-S{\'a}nchez, M., Esteva, M.: A case-based reasoning
  approach for norm adaptation. In: International Conference on Hybrid
  Artificial Intelligence Systems. pp. 168--176. Springer (2010)

\bibitem{conte2001}
Conte, R., Dellarocas, C. (eds.): Social order in multiagent systems. Springer
  (2001)

\bibitem{DomenicoEtAl_2011}
Corapi, D., Russo, A., De~Vos, M., Padget, J., Satoh, K.: Normative design
  using inductive learning. Theory and Practice of Logic Programming
  \textbf{11}(4--5),  783--799 (2011)

\bibitem{DBLP:conf/ecai/CranefieldMOS16}
Cranefield, S., Meneguzzi, F., Oren, N., Savarimuthu, B.T.R.: A {Bayesian}
  approach to norm identification. In: 22nd European Conference on Artificial
  Intelligence. Frontiers in Artificial Intelligence and Applications,
  vol.~285, pp. 622--629. {IOS} Press (2016)

\bibitem{Dam2015}
Dam, H.K., Savarimuthu, B.T.R., Avery, D., Ghose, A.: Mining software
  repositories for social norms. In: 37th International Conference on Software
  Engineering. vol.~2, pp. 627--630. IEEE (2015)

\bibitem{Gao2014}
Gao, X., Singh, M.P.: Extracting normative relationships from business
  contracts. In: International Conference on Autonomous Agents and Multi-Agent
  Systems. pp. 101--108. IFAAMAS (2014)

\bibitem{DBLP:conf/nips/GasthausT10}
Gasthaus, J., Teh, Y.W.: Improvements to the sequence memoizer. In: Advances in
  Neural Information Processing Systems 23, pp. 685--693. Curran Associates,
  Inc. (2010)

\bibitem{GDELTCodeBookV2}
The {GDELT} Event Database --- Data format Codebook v2.0.
  \url{http://data.gdeltproject.org/documentation/GDELT-Event_Codebook-V2.0.pdf}
  (2015)

\bibitem{Kelsen1990}
Kelsen, H.: General theory of norms. Clarendon Press (1990)

\bibitem{Leetaru13gdelt:global}
Leetaru, K., Schrodt, P.A.: {GDELT}: Global data on events, location, and tone,
  1979--2012. In: Proceedings of the International Studies Association Annual
  Convention (2013),
  \url{http://data.gdeltproject.org/documentation/ISA.2013.GDELT.pdf}

\bibitem{Murphy:2012:MLP:2380985}
Murphy, K.P.: Machine Learning: A Probabilistic Perspective. The MIT Press
  (2012)

\bibitem{Qiao2015}
Qiao, F., Li, P., Deng, J., Ding, Z., Wang, H.: Graph-based method for
  detecting occupy protest events using {GDELT} dataset. In: International
  Conference on Cyber-Enabled Distributed Computing and Knowledge Discovery.
  pp. 164--168. IEEE (2015)

\bibitem{Savarimuthu2010}
Savarimuthu, B.T.R., Cranefield, S., Purvis, M.A., Purvis, M.K.: Obligation
  norm identification in agent societies. Journal of Artificial Societies and
  Social Simulation  \textbf{13}(4) (2010),
  \url{https://doi.org/10.18564/jasss.1659}

\bibitem{savarimuthu2013a}
Savarimuthu, B.T.R., Cranefield, S., Purvis, M.A., Purvis, M.K.: Identifying
  prohibition norms in agent societies. Artificial Intelligence and Law
  \textbf{21}(1),  1--46 (2013)

\bibitem{Schrodt2012}
Schrodt, P.A.: {CAMEO: Conflict and Mediation Event Observations Event and
  Actor Codebook} (2012),
  \url{http://data.gdeltproject.org/documentation/CAMEO.Manual.1.1b3.pdf}

\bibitem{DBLP:conf/ijcai/SenA07}
Sen, S., Airiau, S.: Emergence of norms through social learning. In:
  Proceedings of the 20th International Joint Conference on Artificial
  Intelligence. pp. 1507--1512 (2007)

\bibitem{shoham1997}
Shoham, Y., Tennenholtz, M.: On the emergence of social conventions: modeling,
  analysis, and simulations. Artificial Intelligence  \textbf{94}(1),  139--166
  (1997)

\bibitem{DBLP:conf/aaai/TanBS19}
Tan, Z.X., Brawer, J., Scassellati, B.: That's mine! learning ownership
  relations and norms for robots. In: Thirty-Third {AAAI} Conference on
  Artificial Intelligence. pp. 8058--8065. {AAAI} Press (2019)

\bibitem{UlfelderBlog2014}
Ulfelder, J.: Another note on the limitations of event data.
  \url{https://dartthrowingchimp.wordpress.com/2014/06/06/another-note-on-the-limitations-of-event-data}
  (2014)

\bibitem{Wilks1938}
Wilks, S.S.: The large-sample distribution of the likelihood ratio for testing
  composite hypotheses. The Annals of Mathematical Statistics  \textbf{9}(1),
  60--62 (1938)

\bibitem{SequenceMemoizerCACM}
Wood, F., Gasthaus, J., Archambeau, C., James, L., Teh, Y.: The sequence
  memoizer. Communications of the ACM  \textbf{54}(2),  91--98 (2011)

\bibitem{yonamine2013b}
Yonamine, J.E.: A nuanced study of political conflict using the Global Datasets
  of Events Location and Tone ({GDELT}) dataset. Ph.D. thesis, Pennsylvania
  State University (2013)

\end{thebibliography}
%
%
%
%
%
\end{document}